\documentclass{article} 
\usepackage{iclr2023_workshop,times}

\usepackage{amsmath,amsfonts,bm}

\def\eqref#1{equation~\ref{#1}}

\def\1{\bm{1}}

\DeclareMathAlphabet{\mathsfit}{\encodingdefault}{\sfdefault}{m}{sl}
\SetMathAlphabet{\mathsfit}{bold}{\encodingdefault}{\sfdefault}{bx}{n}

\DeclareMathOperator*{\argmin}{arg\,min}

\usepackage{hyperref}
\usepackage{url}

\usepackage[utf8]{inputenc} 
\usepackage[T1]{fontenc}    
\usepackage{hyperref}       
\usepackage{url}            
\usepackage{booktabs}       
\usepackage{amsfonts}       
\usepackage{nicefrac}       
\usepackage{microtype}      
\usepackage{lipsum}		
\usepackage{graphicx}
\usepackage{doi}
\usepackage{soul}

\usepackage[utf8]{inputenc} 
\usepackage[T1]{fontenc}    
\usepackage{hyperref}       
\usepackage{url}            
\usepackage{booktabs}       
\usepackage{amsfonts}       
\usepackage{nicefrac}       
\usepackage{microtype}      

\usepackage{amsmath}
\usepackage{graphicx}
\usepackage[export]{adjustbox}
\usepackage{dsfont}
\usepackage{booktabs}
\usepackage{multirow}
\usepackage{wrapfig}
\usepackage{float}
\usepackage{subfigure}
\usepackage{ownstyles}
\usepackage{caption}
\usepackage{tabularx}
\usepackage{chngcntr}

\title{Nature's Cost Function: Simulating Physics by Minimizing the Action}

\author{Tim Strang \\
ML Collective \\
Holliston, MA, 01746 \\
\texttt{timdstrang@gmail.com} \\
\And
Isabella Caruso \\
MIT Materials Science \\ 
Cambridge, MA, 02142 \\
\texttt{icaruso@mit.edu} \\
\And
Sam Greydanus \\
ML Collective \\
Corvallis, Oregon, 97330 \\
\texttt{greydanus.17@gmail.com} \\
}

\iclrfinalcopy
\begin{document}

\maketitle

\begin{abstract}
In physics, there is a scalar function called the action which behaves like a cost function. When minimized, it yields the ``path of least action'' which represents the path a physical system will take through space and time. This function is crucial in theoretical physics and is usually minimized analytically to obtain equations of motion for various problems. In this paper, we propose a different approach: instead of minimizing the action analytically, we discretize it and then minimize it directly with gradient descent. We use this approach to obtain dynamics for six different physical systems and show that they are nearly identical to ground-truth dynamics. We discuss failure modes such as the unconstrained energy effect and show how to address them. Finally, we use the discretized action to construct a simple but novel quantum simulation. Code: \texttt{github.com/greydanus/ncf}
\end{abstract}

\begin{figure}[h!] \centering
\includegraphics[width=\textwidth]{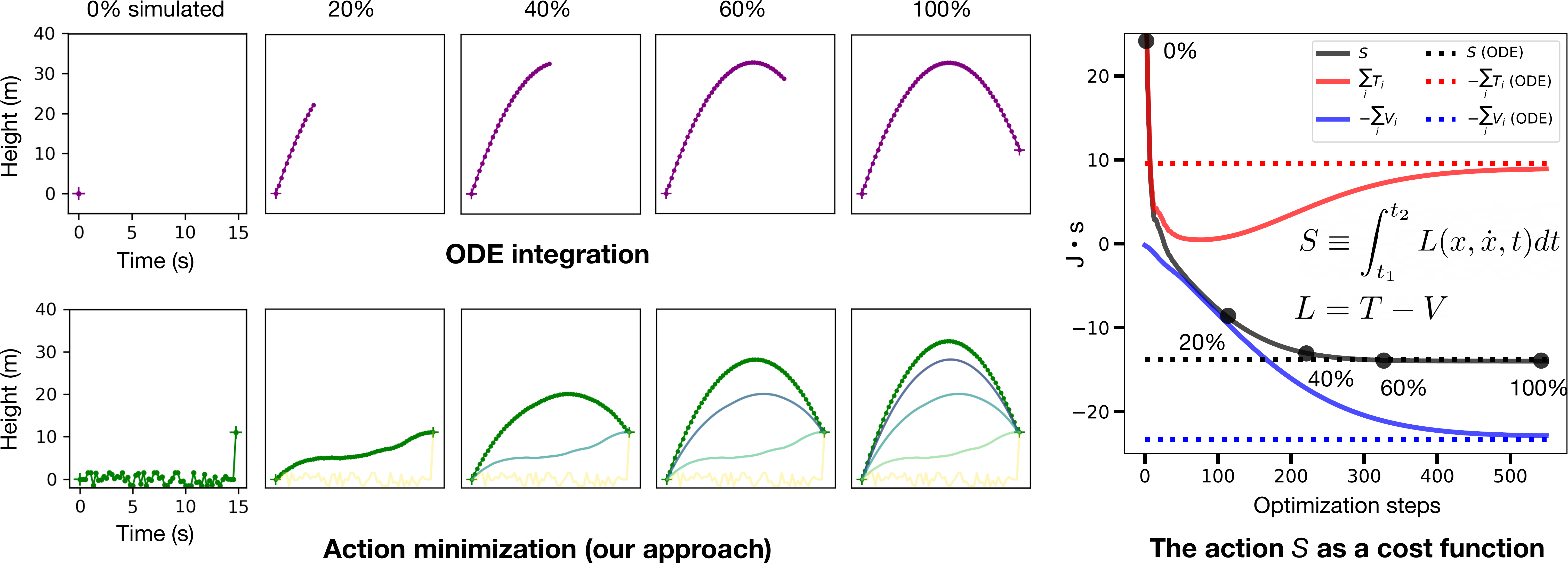}
\caption{Finding a path of least action with gradient descent. \textbf{Left}: We compare the normal approach of ODE integration to our approach of action minimization. The action is the sum, over every point in the path, of kinetic energy $T$ minus potential energy $V$. We compute the gradients of this quantity with respect to the path coordinates and then deform the initial path (yellow) into the path of least action (green). This path resolves to a parabola, matching the path obtained via ODE integration. \textbf{Right}: We plot the path’s action $S$, kinetic energy $T$, and potential energy $V$ over the course of optimization. All three quantities asymptote at the $S$, $T$, and $V$ values of the ODE trajectory.}
\label{fig:fig1}
\end{figure}

\section{Introduction}

Over the past two centuries, many works have studied the action in a variety of physical and mathematical settings \citep{goldstine2012history, ferguson2004brief}. In all these studies, the authors treated the action analytically. They only solved for paths of least action indirectly by first invoking the Euler-Lagrange equation, then solving for a system of differential equations, and finally integrating those equations over time with an ordinary differential equation (ODE) solver.

In this paper, we take a different approach:
\begin{quote}\textit{Instead of solving for paths of stationary action analytically, we discretize the action function and then minimize it directly with gradient descent.}\end{quote}
We use this approach to obtain dynamics for six different physical systems: a body in free fall, a pendulum, a double pendulum, the three body problem, a gas with a Lennard-Jones potential, and the planetary ephemerides of the inner solar system. We document a failure mode called the \textit{unconstrained energy effect} and discuss how to mitigate it. Finally, we note that the action controls the resonance modes of paths in quantum mechanics and use our discretized version to construct a simple quantum simulation.

\section{Background}

\textbf{The Lagrangian method.} The Lagrangian method begins by considering all the paths a physical system could take from an initial state $\bf x(t_0)$ to a final state $\bf x(t_1)$. Then it provides a simple rule for selecting the path $\hat{\bf x}$ that nature will actually take: the action $S$, defined in Eqn. \ref{eqn:the-action}, must have a stationary value over this path. Here $T$ and $V$ are the kinetic and potential energy functions for the system at any given time $t$ in $[t_0,t_1]$.
\begin{align}
\label{eqn:the-action}
S &:= \int_{t_0}^{t_1} L({\bf x}, ~ \dot{\bf x}, ~ t) ~ dt
\quad \textrm{where}\quad L = T - V \\
\label{eqn:euler-lagrange}
\quad \hat{\bf x} &~~ \textrm{has the property} \quad \frac{d}{dt} \left( \frac{\partial L}{\partial \dot{\hat{x}}(t)} \right) = \frac{\partial L}{\partial \hat{x}(t)} \quad \textrm{for} \quad t \in [t_0,t_1]
\end{align}
\textbf{Finding $\hat{\bf x}$ with Euler-Lagrange.} When $S$ is stationary, we can show that the Euler-Lagrange equation (Eqn. \ref{eqn:euler-lagrange}) holds true over the interval $[t_0,t_1]$ \citep{morin2008introduction}. This observation is valuable because it allows us to solve for $\hat{\bf x}$: first we apply the Euler-Lagrange equation to the Lagrangian $L$ and derive a system of partial differential equations. Then we integrate those equations to obtain $\hat{\bf x}$. Importantly, this approach works for all problems spanning classical mechanics, electrodynamics, thermodynamics, and relativity. It provides a coherent theoretical framework for studying classical physics as a whole.

\textbf{Finding $\hat{\bf x}$ with action minimization (this work).} A more direct approach to finding $\hat{\bf x}$ begins with the insight that paths of stationary action are almost always \textit{also} paths of least action \cite{morin2008introduction}. Thus, without much loss of generality, we can exchange the Euler-Lagrange equation for the simple minimization objective shown in the third part of Eqn. \ref{eqn:discretization}. Meanwhile, as shown in the first part of Eqn. \ref{eqn:discretization}, we can redefine $S$ as a discrete sum over $N$ evenly-spaced time slices:
\begin{align}
\label{eqn:discretization}
S := \sum_{i=0}^{N} L({\bf x}, ~ \dot{{\bf x}}, ~ t_i) \Delta t \quad \textrm{where} \quad \dot{{\bf x}}(t_i) := \frac{ {\bf x}(t_{i+1}) - {\bf x}(t_{i})}{\Delta t} \quad \textrm{and} \quad \hat{\bf x} := \argmin_{\bf x} S(\bf x)
\end{align}
One problem remains: having discretized $\hat{ \bf x}$ we can no longer take its derivative to obtain an exact value for $\dot{ \bf x}(t_i)$. Instead, we must use the finite-differences approximation shown in the second part of Equation \ref{eqn:discretization}. Of course, this approximation will not be possible for the very last $\dot{ \bf x}$ in the sum because $\dot{ \bf x}_{N+1}$ does not exist. For this value we will assume that, for large $N$, the change in velocity over the interval $\Delta t$ is small and thus let $\dot{ \bf x}_N = \dot{ \bf x}_{N-1}$. Having made this last approximation, we can now compute the gradient $\frac{\partial S}{\partial {\bf x}}$ numerically and use it to minimize $S$. This can be done with PyTorch \citep{paszke2019pytorch} or any other package that supports automatic differentiation.

\textbf{Related work: minimizing the Onsager-Machlup action.} One line of related work involves minimizing the action of the Onsager-Machlup (OM) function \citep{onsager1953fluctuations, faccioli2006dominant}. It is used to describe the time dynamics of the probability density of a stochastic process. The action of the OM function has a different physical interpretation from the action of a Lagrangian but shares certain mathematical properties. \cite{adib2008stochastic} discretize diffusion trajectories and calculate their gradients (Eqn. 22) but do not minimize the OM action. Conversely, \cite{lee2017finding, zuckerman2000efficient} minimize the OM action -- to analyze protein folding dynamics -- but use gradient-free optimization methods: Metropolis MCMC and Conformational Space Annealing respectively \citep{metropolis1953equation,lee1997new}. Like our work, these works fix the initial and final states and (sometimes) minimize the action over the interval. Unlike our work, they use the OM function and do not perform minimization with automatic differentiation or gradient descent. We discuss additional related work in the context of gradient-based optimization of physics problems in Appendix \ref{appx:related}.

\begin{figure}[t!]
\vspace{-1cm}
\centering
\includegraphics[width=\textwidth]{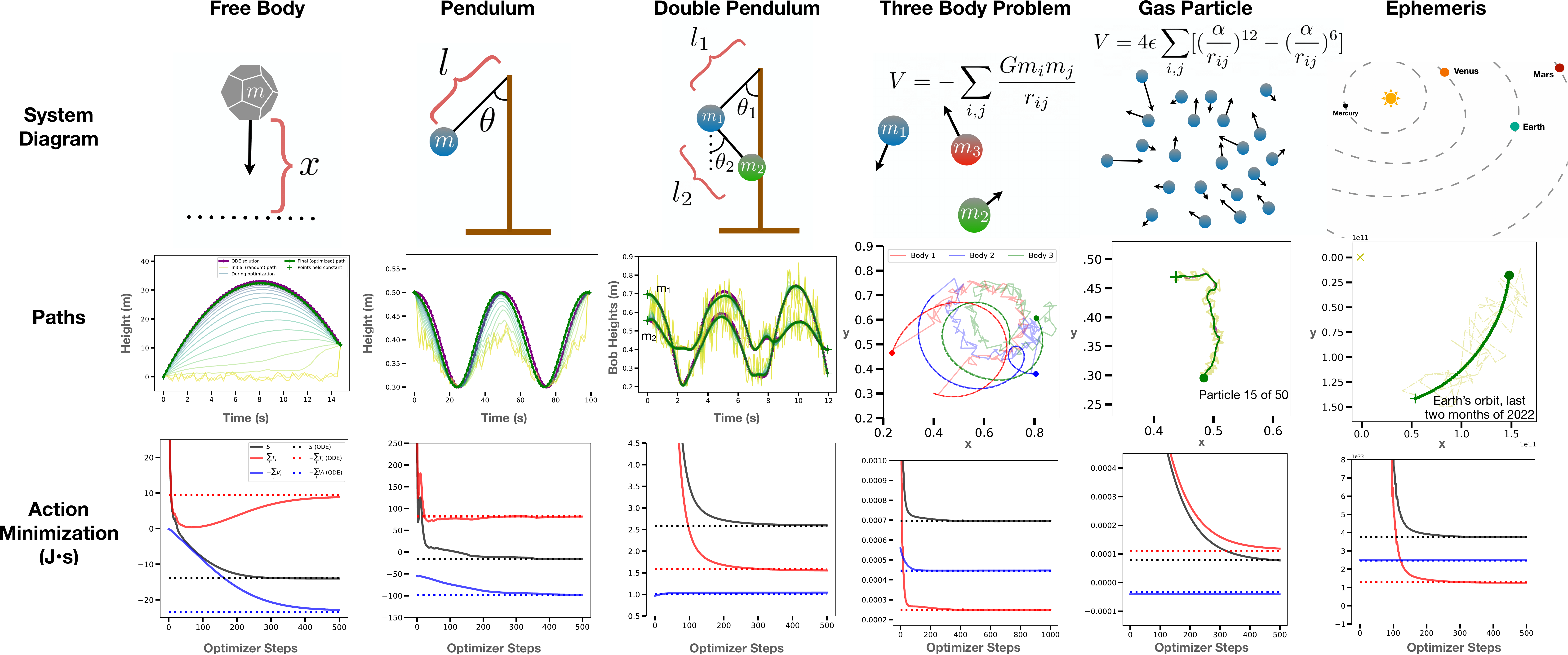}
\caption{Visualizing our technique applied to a variety of different physics simulations. \textbf{First row}: We show diagrams of the setups and variables of the six systems under investigation. \textbf{Second row}: we visualize the ODE paths (purple), initial semi-random paths (yellow), and final paths of minimized action (green). These plots give an intuition for how the paths deform during optimization. \textbf{Third row}: We visualize the optimization dynamics in terms of action, kinetic energy, and potential energy. }
\label{fig:fig2}
\end{figure}

\begin{wrapfigure}{r}{0.6\textwidth}
  \vspace{-1.9cm}%
  \begin{center}
    \includegraphics[width=0.55\textwidth]{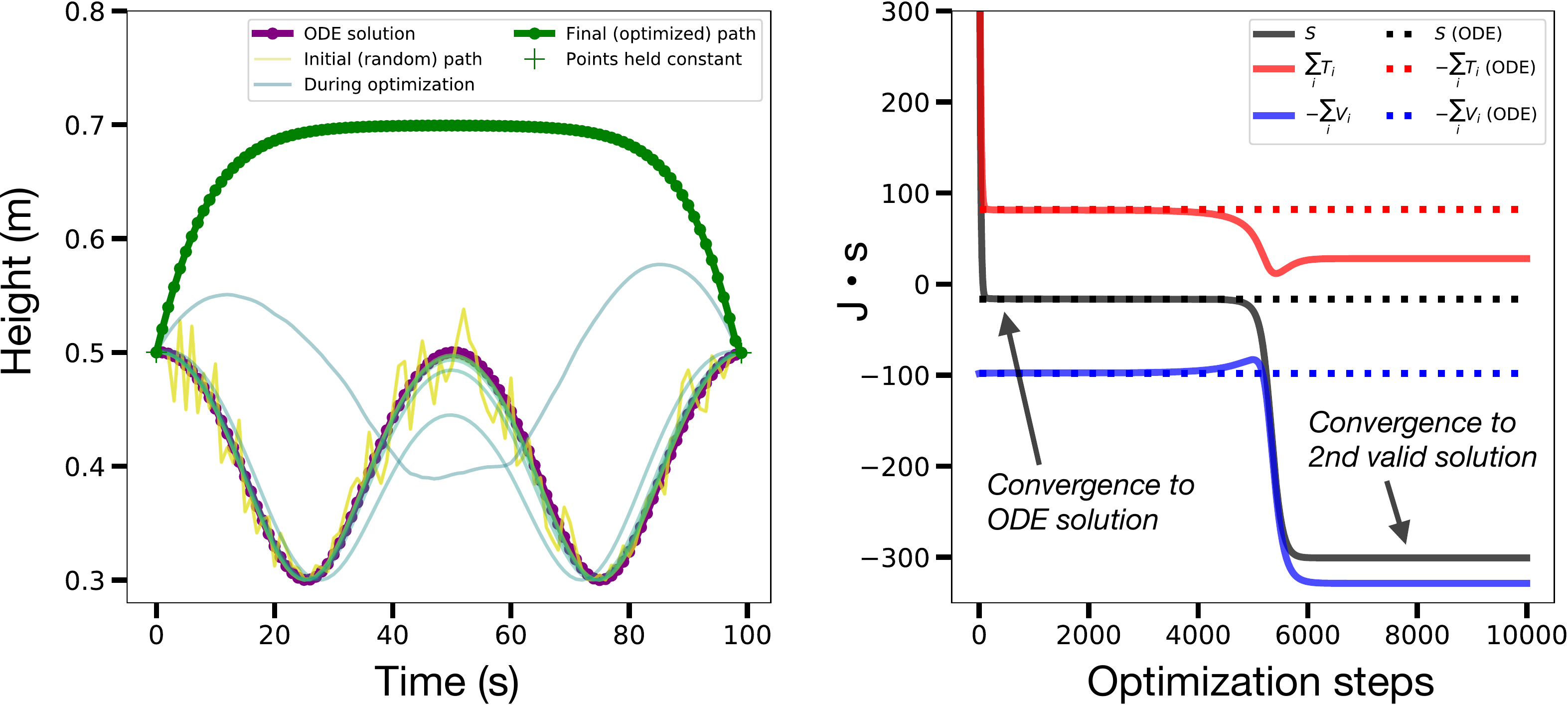}
  \end{center}
    \caption{The \textit{unconstrained energy effect} in the pendulum task. Optimization converges to the baseline path initially. After additional optimizer steps, it converges to a different but also valid solution where the pendulum swings up, hangs vertically, and then falls into the final state. This occurs because our current methods do not constrain the total energy $T+V$ in the way the ODE initial state $({\bf x}(t_0), \dot{{\bf x}}(t_0))$ does.}
  \label{fig:fig3}
  \vspace{-0.25cm}
\end{wrapfigure}

\section{Six Experiments}
\label{sec:methods}

Minimizing the action in the way we have described has not been studied in detail. We have discussed a few related works which do this in the context of the OM function but their scope and details diverge from this work and each other. Thus in our experiments we prioritized simplicity. Unless otherwise specified, we set all constants such as mass and gravity to one. When selecting physical systems, we began with two toy problems (for debugging): a free body and a pendulum. Then we investigated four more complex systems: a double pendulum, the three body problem, a simple gas, and a real ephemeris dataset of planetary motion. These systems presented an interesting challenge because they were all nonlinear, chaotic, and high-dimensional\footnote{The state of the simple gas, for example, has a hundred degrees of freedom.}. In each case, we compared our results to a baseline path obtained with a simple ODE solver using Euler integration. Appendix \ref{appx:methods} gives more details, including the Lagrangians and equations of motion.

\textbf{The unconstrained energy effect.} Early in our experiments we encountered \textit{the unconstrained energy effect}. This happens when the optimizer converges on a valid physical path with a different total energy from the baseline. Figure \ref{fig:fig3} shows an example. The reason this happens is that, although we fix the initial and final states, we do not constrain the path's total energy $T+V$. Even though paths like the one in Figure \ref{fig:fig3} are not necessarily invalid, they make it difficult for us to recover baseline ODE paths. For this reason, we use the baseline ODE paths to initialize our paths, perturb them with Gaussian noise, and then use early stopping to select for paths which are similar (often, identical) to the ODE baselines. This approach matches the mathematical ansatz of the ``calculus of variations'' where one studies perturbed paths in the vicinity of the true path of least action. We note that there are other ways to mitigate this effect which don't require an ODE-generated initial path. We discuss them in Appendix \ref{appx:methods}, as they are beyond the main scope of this work.

\textbf{Results.} On all six physical systems we obtained paths of least action which were nearly identical to the baseline paths of the ODE solver. Figure \ref{fig:fig2} shows optimization dynamics and qualitative results while Table \ref{tab:tab2} shows quantitative results. These results suggest that action minimization can generate physically-valid dynamics even for chaotic and strongly-coupled systems like the double pendulum and three body problem. One interesting pattern we noticed was that optimization dynamics were dominated by the kinetic energy term $T$ (third row of Figure \ref{fig:fig2}). This occurs because $S$ tends to be more sensitive to $T$ (which grows as $\dot{{\bf x}}^2$) than $V$. Future methods should focus on stabilizing $T$.

\section{The Quantum Limit}

\begin{wrapfigure}{H}{0.5\textwidth}
  \vspace{-2cm}%
  \begin{center}
    \includegraphics[width=0.5\textwidth]{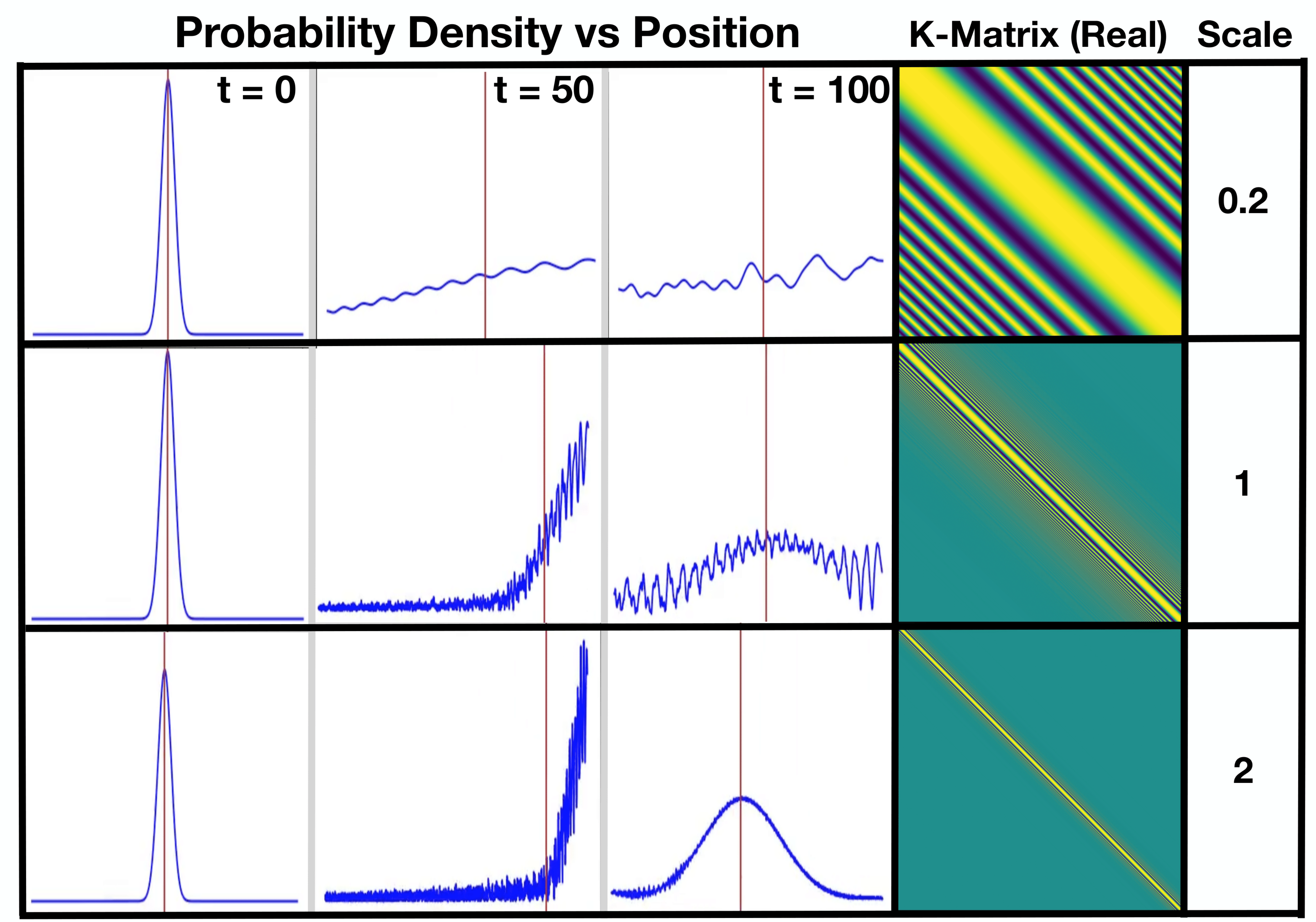}
  \end{center}
  \caption{Visualizing wave packet dynamics at different scales ($\hbar$ $\propto$ Scale$^{-1}$). As $\hbar\rightarrow0$, the waves retain coherence longer and interfere with the right wall at higher frequencies, approaching the behavior of a classical particle at Scale = 2. Coherence grows as oscillations along the off-diagonals of the phase-action matrix, $K$, increase in frequency.}
  \label{fig:fig4}
  \vspace{-0.25cm}
\end{wrapfigure}

The equation for the action, which Lagrange introduced in the context of classical mechanics, is one of the few equations which does not need to be redefined in the quantum context. Here, systems are described with a wave function and allowed to take on a superposition (linear combination) of paths at once. The best-known example of this phenomena is the famous double-slit experiment, where a single electron fired at two separate slits travels through both and interferes with itself on the other side. \textit{This interference occurs because the phase evolves differently with each path and thus is out of sync when the paths recombine.} The exponential portion of Eqn. \ref{eqn:path-integral} describes how a system evolves as it travels along a given path: it oscillates at a rate proportional to the path's action. Thus the \textit{phase} of the wave function, after traveling along a given path for some interval of time, is also proportional to the action. The rest of Eqn. \ref{eqn:path-integral} describes how to recombine the contributions of all paths after some interval of time. First, it rotates the wave function by the phase change associated with each path and then it sums (integrates) these states to obtain the full state at the end of the time interval. In the double slit example, this summation would recombine the waves from the two slits into the full wave function $\Psi(\bf x, t)$ sampled at the sensor.
\begin{align}
 \Psi(x, t) = \int\limits_{\bf x(0)=x}  \Psi(\bf x(t), 0) * exp\{\frac{i}{\hbar}\int L[\bf x(\tau),\; \bf \dot x(\tau), \tau]d\tau\}\mathcal \; \mathcal D\bf x
 \label{eqn:path-integral}
 \end{align}
 \textbf{Discretizing the path integral.} \cite{feynman1965feynman} and many others have treated Eqn. \ref{eqn:path-integral} analytically, deriving from it both Schr\"odinger's equation and the classical principle of least action. However, as was the case with the classical action, few works have discretized the paths and used the action to construct a numerical simulation. Once again, that is our goal. It can be done by dividing the spatial and time dimensions of Eqn. \ref{eqn:path-integral} into small, uniform intervals. Then, for each time interval, we can restrict our group of paths to only the $N^2$ linear paths that connect the $N$ discrete spatial locations to one another. We can calculate their actions and, with the help of Eqn. \ref{eqn:path-integral}, their phase changes. Then we can organize these phase changes in an $N \times N$ \textit{phase-action} matrix, $K$, where rows correspond to locations of path start points, columns correspond to locations of path endpoints, and the matrix elements themselves are the phase changes of the paths. This matrix, plotted in Figure \ref{fig:fig4}, provides an intuitive way to visualize how nearby paths interfere with one another at the quantum-classical limit.
 
\textbf{A simple quantum simulation.} Now, to simulate dynamics over one discrete time slice, we can simply multiply the wave function (a discretized vector of complex amplitudes at the $N$ spatial locations) by $K$. Then we can evolve it arbitrarily far into the future by multiplying by $K$ repeatedly (see Figure \ref{fig:fig4}). The quantum simulation we recover is fundamentally different from the more popular Hamiltonian-based simulations \citep{hartree1935self, richings2015quantum} and is remarkable for its simplicity. It mirrors the philosophical approach we took in the six classical experiments in that, in both cases, we express computation in terms of the action as directly as possible.

\section{Conclusion}

Basic physical principles and the elegant reasoning behind them are often obscured in the midst of numerical approximations and domain-specific notation. In this work we have sought to prevent this from happening by expressing computation in terms of the action as directly as possible.




\subsubsection*{Acknowledgments}
All authors thank Jason Yosinski, Rosanne Liu, and other members of the ML Collective research group for helpful discussion, ideas, and feedback on experiments.

TS thanks the Central Rock Gym for an ambient environment encouraging of intellectual reflection. IC thanks the MIT Materials Science Department for unwittingly allowing her to indulge in this side project while waiting for feedback on her thesis application. SG thanks TS and IC for joining him on this research journey which was neither the shortest path nor the one of least action.

\bibliography{iclr2023_workshop}
\bibliographystyle{iclr2023_workshop}

\newpage

\appendix
\section{Appendix}

\subsection{Additional background}
\label{appx:related}

\textbf{History and interpretation of the action}. Physicists discovered the action in the $19^{\textrm{th}}$ century while attempting to derive the mechanics of solid and fluid bodies on a purely analytic basis \cite{ferguson2004brief}. There was a desire to build a coherent, mathematically-consistent framework for describing the physical world. Notions of physics at the time were highly intuitive and lacked rigorous analytic treatment. Thus at its conception the utility of the action was the main focus, not its physical interpretation. Even so, the interpretation of the action is of fundamental interest in physics. For the most part (there are exceptions in relativity) the action is written as the difference between the summed kinetic energy and the summed potential energy along a (closed) system's path through time (see Eqn. \ref{eqn:the-action}. Our intuition is that these two terms roughly correspond to the total flow of energy due to entropic forces and inertial forces respectively. Paths of least action, then, represent the notion that the dynamics of the universe are the result of a delicate balance between forces associated with entropy and forces associated with inertia.

\textbf{The Newtonian vs. Lagrangian view of physics}. When studying physics, one begins with concepts that are most accessible to human intuition and then proceeds towards more abstract concepts that describe nature in generality. This, for example, is why introductory physics courses emphasize a Newtonian view of physics. Here a group of objects are studied at a particular snapshot in time, their forces tabulated, and their positions updated to yield the next state of the system. This view is intuitive because it matches our natural experience of time and space. The Lagrangian method, by contrast, treats the entire path of a physical system as its own mathematical entity. It is the more abstract and unintuitive approach, but it is also far more general and mathematically consistent across domains.

\textbf{Related work: gradient-based optimization for physics.} We have already addressed how other works studied minimization of the OM action without using gradient-based methods. There are also a number of works that have explored the use of automatic differentiation and gradient-based optimization in the physical sciences, but have not specifically looked at the action. \cite{schoenholz2020jax} use these tools to study phenomena such as phase transitions and crystal packing in the context of molecular dynamics. \cite{freeman2021brax} introduce a general-purpose differentiable physics engine for research in policy optimization. Another recent line of work combines gradient-based optimization with differential equations (DE). For example, \cite{chen2018neural} proposed using neural networks to parameterize ODEs and \cite{lutter2019deep, cranmer2020lagrangian} showed how to do the same for Lagrangian DEs. Meanwhile, work by \cite{bar2019learning} and others has focused on improving adaptive integration methods.

\subsection{Additional methods}
\label{appx:methods}

\textbf{Optimization.} We experimented with standard gradient descent, gradient descent with momentum, and the Adam optimizer \cite{kingma2014adam}, which adaptively scales momentum. Adding momentum -- either directly or via Adam -- improved convergence time by at least a factor of two. We chose the Adam optimizer so as to avoid setting separate momentum hyperparameters. We used 500 optimization steps for all of our experiments except for the three body problem, which recovered, with near-exact precision, the ODE path after 1000 steps.

\textbf{Constructing the baseline paths.} To simulate a classical system's dynamics using traditional methods, we must first derive a set of second-order differential equations of motion from physical principles. This is generally done via application of Newton's Second Law or the Euler-Lagrange equation (Eqn. \ref{eqn:euler-lagrange}). In all systems discussed in this paper, these equations of motion are directly derivable and expressed as sets of coupled ordinary differential equations (ODEs) in Table \ref{tab:tab1}. In order to set up a traditional baseline for contrasting with the new action optimization method, each of these was solved numerically using Euler's method prior to optimizer trials, using boundary conditions known to produce complex behaviors. Euler's method is the most basic and broadly understood numerical ODE-solving technique, and  was chosen so as to maximize the baseline's analogy to our equally basic but as-yet untested optimization-based method.

\textbf{Other ways to address the unconstrained energy effect.} 
Several alternative methods were implemented in attempts to address the unconstrained energy effect before the path perturbation method discussed in the main body of the paper was finally selected. Initially, upon realizing the problem lay in a lack of sufficient constraints, we attempted to feed the optimizer the same ${\bf x}(t_0)$, $\dot{{\bf x}}(t_0)$ information that the ODE solver used to resolve total system energy. The simplest way to do this in our discretized path model was to give the optimizer more reference points from the baseline ODE simulation. Practically, this meant freezing not just the initial and final points of the path during optimization, but several of the adjacent points as well. While this did constrain the initial and final velocities of the optimized paths, it did not significantly affect the behavior of the entire path unless large swathes of points were frozen. As this would greatly reduce the space of potential applications, we shifted our attention to other possible solutions. 

Next, two distinct regularization schemes were implemented, both based on the addition of energy conservation terms in the optimizer loss function. The first added a global energy loss term, which penalized the optimizer in proportion to the difference between total system energy at each point along a path and the expected value, calculated from ODE ${\bf x}(t_0)$ and $\dot{{\bf x}}(t_0)$ input conditions. The second regularizer did not rely on the ODE's initial conditions, but added a loss term proportional to the derivative of total system energy over time, i.e. a local energy conservation term. The first of these worked to a degree, but the path perturbation method gave far cleaner results on the majority of systems tested. Additionally, path perturbation showcases the potential ability of the optimizer method to refine energy predictions based on coarser models. Especially in chemistry, precise energy predictions are a common target for computational models. The local energy conservation method, on the other hand, provided no noticeable advantage. Indeed, it sometimes had an overtly adverse effect on optimizer performance, likely due to the additional restrictions it enforced on the space of potential intermediate paths.

For the purposes of our exploratory work, the path perturbation method was the best fit. We suspect that there is a better method for properly constraining the total energy and pointing action minimization towards one non-degenerate path. One of the main focuses of future work in this area should be the characterization of such a method.

\begin{table}[H]
\begin{center}
\begin{tabular}{l|lllll}
\multicolumn{1}{l}{System} &\multicolumn{1}{l}{Lagrangian $\mathcal{L} = T-V $ } & \multicolumn{1}{l}{Equations of motion} &\multicolumn{1}{l}{Noise $\sigma$}  &\multicolumn{1}{l}{\bf $\Delta t$} &\multicolumn{1}{l}{LR}
\\ \hline \\ 

Free Body &
$\frac{1}{2} m\dot{x}^2-mgx$ & $\ddot x = g$  & $1.5$ & $0.25$ & $1$  \\

\\ Pendulum &
$\frac{1}{2}ml^2\dot\theta^2 - mgl(1-cos(\theta))$ &$\ddot\theta = -\frac{g}{l}sin(\theta)$ &$2e$-$1$ & $1$ & $5e$-$2$  \\

\\ Dbl. Pend. &
 Appendix &Appendix &$6e$-$1$ & $6e$-$2$ & $1e$-$2$  \\
 
 \\ Three Body &
$\sum\limits_{i,j}\frac{1}{2}m_i\bf\dot x_i^2 + \frac{Gm_im_j}{|r_{ij}|}$  &$\bf\ddot x_i = -G\sum\limits_j \frac{m_j}{|r_{ij}|^2}\hat r_{ij}$ &$3e$-$2$ & $0.5$ &  $2e$-$4$ \\

\\ Gas &
 $\sum\limits_{i,j}\frac{1}{2}m_i\bf\dot x_i^2 + V_{LJ}(\bf r_{ij})$ &$\bf\ddot x_i = \frac{1}{m_i} \sum\limits_j F_{LJ}(\bf r_{ij})$ & $1e$-$2$ & $0.5$ &  $1e$-$4$  \\
 
 \\ Ephemeris &
$\sum\limits_{i,j}\frac{1}{2}m_i\bf\dot x_i^2 + \frac{Gm_im_j}{r_{ij}}$ &$\bf\ddot x_i = -G\sum\limits_j \frac{m_j}{r_{ij}^2}\hat r_{ij}$  & $2e$+$10$ & $1$ day & $1e$+$9$   \\ \hline

\end{tabular}
\end{center}
\caption{All optimizations were run for 500 steps. LR = learning rate.}
\label{tab:tab1}
\end{table}

Double pendulum Lagrangian:
\[ \bf L = \frac{1}{2}(m_1+m_2)l_1^2\dot\theta_1^2 \;+\; 
    \frac{1}{2}m_2l_2^2\dot\theta_2^2 \;+\;  m_2l_1l_2\dot\theta_1\dot\theta_2cos(\theta_1-\theta_2) \;+\;
    (m_1+m_2)gl_1cos(\theta_1) \;+\; m_2gl_2cos(\theta_2)
\]
Double pendulum equations of motion:
\[ \ddot \theta_i = \frac{f_i-\alpha_if_j}{1-\alpha_i\alpha_j}
 \quad \textrm{where} \quad
 f_1 \equiv -\frac{l_2}{l_1}[\frac{m_2}{m_1+m_2}\dot\theta_2^2sin(\theta_1-\theta_2) - \frac{g}{l_1}sin(\theta_1)]
\]
\[ f_2 \equiv \frac{l_1}{l_2}\dot\theta_1^2sin(\theta_1-\theta_2) - \frac{g}{l_2}sin(\theta_2)
\]
Lennard-Jones Potential:
\[V_{LJ}(\bf r)=4\epsilon \left [ \left ( \frac{\sigma }{r} \right )^{12}-\left (\frac{\sigma }{r}  \right )^{6} \right]\]
where r is the interparticle distance, $\epsilon$ is the depth of the potential energy well in the interaction between the particles, and $\sigma$ is related to the value of r where the potential energy is at the minimum. Using the formula $F = -\nabla U$, we can also derive a Lennard-Jones force:
\[F_{LJ}(\bf r)=4\epsilon \left [ \left ( \frac{12\sigma^{12}}{r^{13}} \right )-\left (\frac{6\sigma^{6}}{r^7}  \right ) \right]\]

\begin{table}[H]
\begin{center}
\begin{tabular}{l|l|l|l|l|l|l|l|l|l}

\multicolumn{1}{l|}{} &\multicolumn{2}{c|}{Sum Action S (Js)} & \multicolumn{2}{c|}{Sum Kinetic T (Js)} &\multicolumn{2}{c|}{Sum Potential V (Js)}  &\multicolumn{2}{c|}{Mean Square Error} \\ \hline \\[-1em]
 System&ODE&Sim&ODE&Sim&ODE&Sim&Initial&Final\\ \hline \\[-1em]
 
Free Body & -13.8 & -14.0 & 9.54 & 8.84 & 23.4 & 22.8 & 630 & 0.352 \\

Pendulum & -16.3 & -16.3 & 81.9 & 81.7 & 98.2 & 98.0 & 0.118 & 1.20e-2 \\

Dbl Pend & 2.59 & 2.59 & 1.58 & 1.55 & -1.01 & -1.04 &0.170 & 3.22e-3 \\

3 Body & 6.94e-4 & 6.96e-4 & 2.48e-4 & 2.49e-4 & -4.46e-4 & -4.47e-4 & 9.74e-3 & 5.74e-6 \\

Gas & 7.81e-5 & 7.71e-5 & 1.11e-4 & 1.18e-4 & 3.30e-5 & 4.13e-5 & 3.09e-5 & 2.80e-6 \\

Ephemeris & 3.76e33 & 3.75e33 & 1.28e33 & 1.26e33 & -2.48e33 & -2.49e33 & 1.50e20 & 7.12e17 \\ \hline

\end{tabular}
\end{center}
\caption{Quantitative results for the six classical systems described in Section \ref{sec:methods}.}
\label{tab:tab2}
\end{table}

\begin{wrapfigure}{r}{0.5\textwidth}
  \vspace{-1.25cm}%
  \begin{center}
    \includegraphics[width=0.5\textwidth]{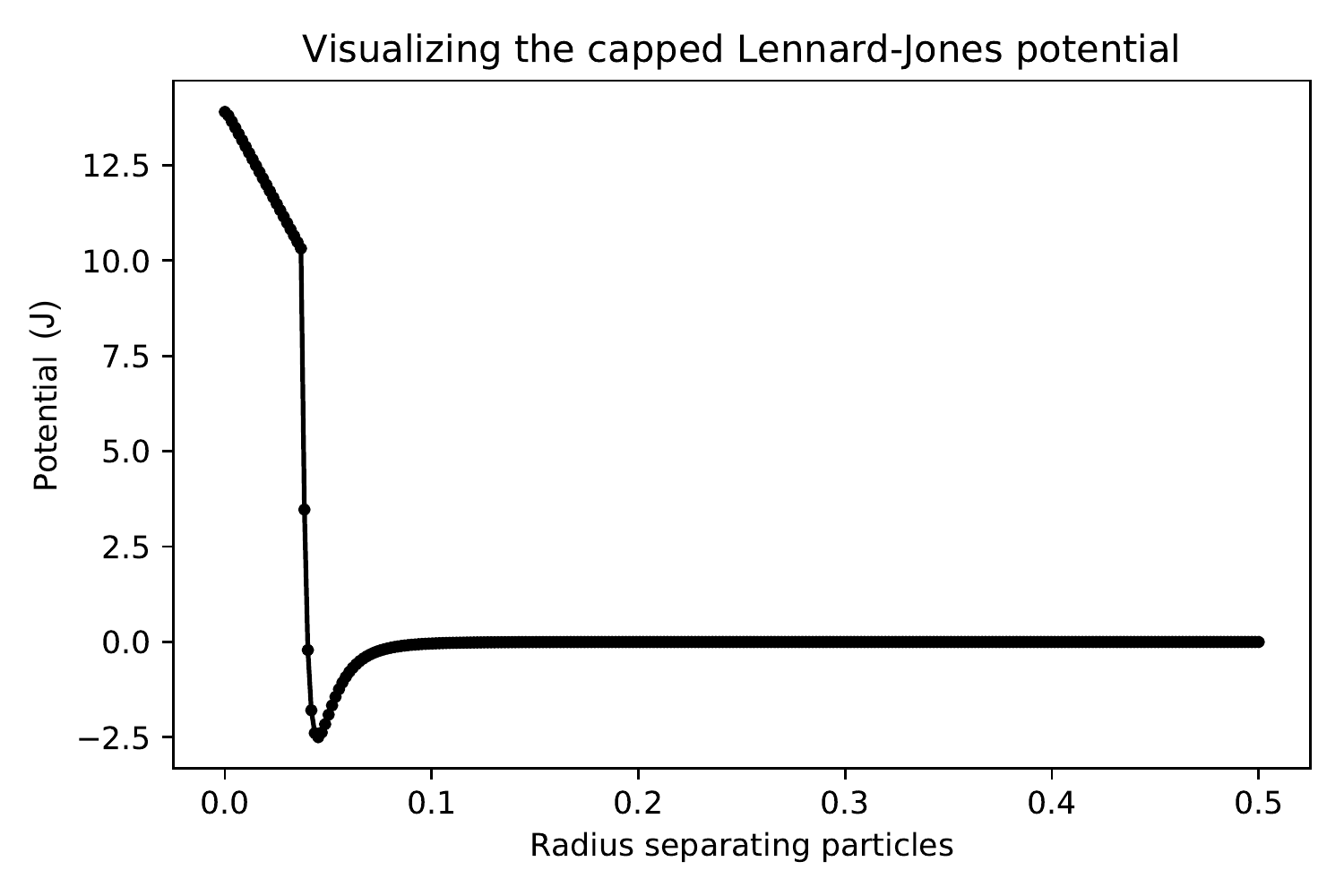}
  \end{center}
    \caption{Visualizing the capped Lennard-Jones potential. The capped portion of the potential has a linearized slope to prevent forces associated with small radii from growing extremely large and creating numerical instabilities.}
  \label{fig:capped-lj}

  \begin{center}
    \includegraphics[width=0.5\textwidth]{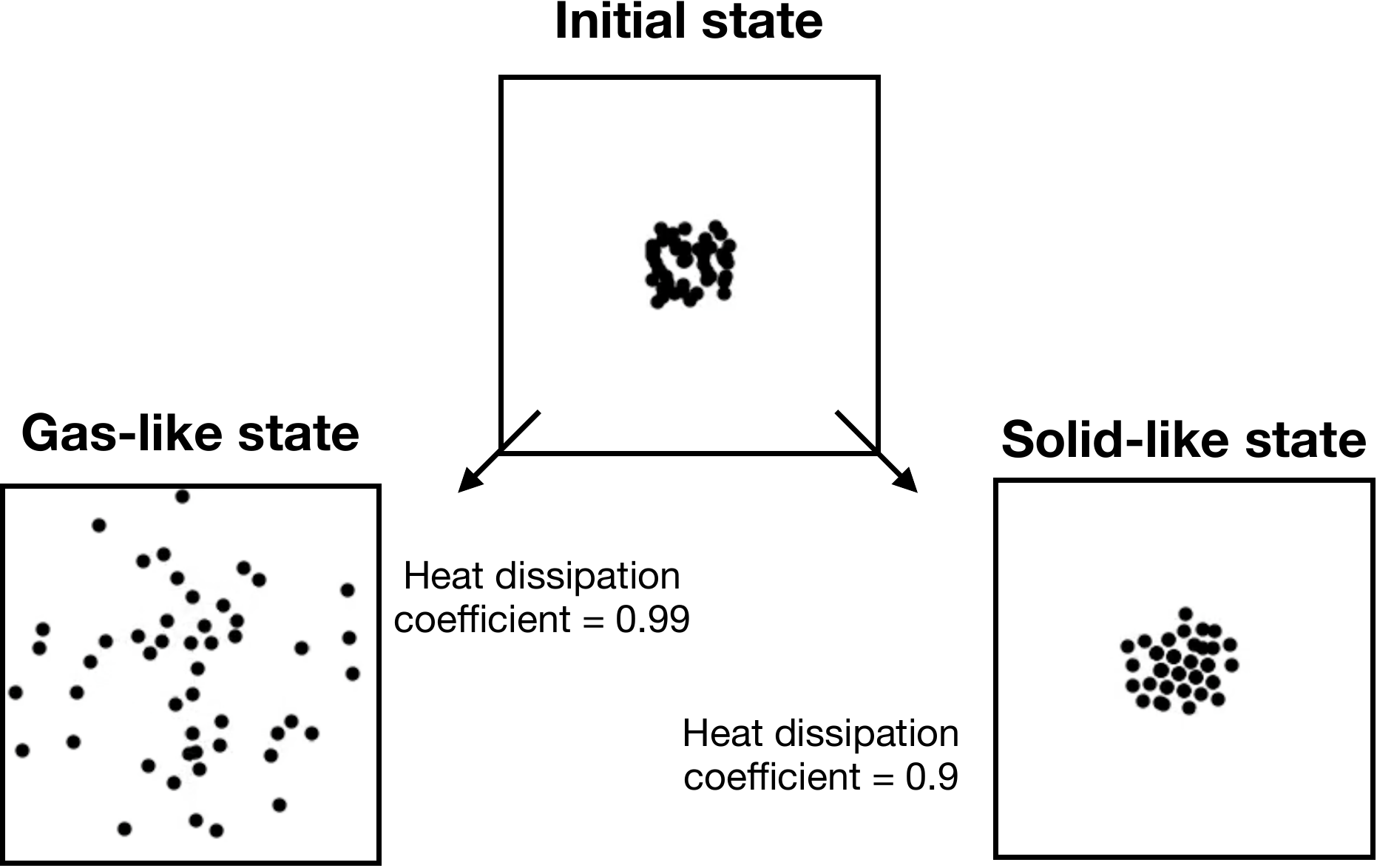}
  \end{center}
    \caption{Visualizing solid and gas states generated by using two different heat dissipation coefficients.}
  \label{fig:states-lj}
  \vspace{-.7cm}
\end{wrapfigure}

\textbf{Gas simulation.} Molecular dynamics (MD) simulations enable investigation of dynamic systems that would be difficult to observe experimentally \citep{hollingsworth2018molecular}. These types of simulations date back to the 1950s, with the earliest work studying hard sphere gasses \citep{alder1959studies}. Today, MD is used to understand far more complicated systems, from helium diffusion in titanium to protein folding \citep{zhang2013molecular, lee2017finding}. These lines of work have advanced in step with our computing abilities. At their most basic and traditional, MD simulations operate by taking the locations of all atoms in the system, using some potential to model interactions between the atoms, calculating each pairwise interaction, and updating the velocities and positions of the atoms accordingly \citep{schroeder2015interactive}.

Our simulation is similar to early MD simulations and simple MD simulations such as \cite{schroeder2015interactive} that are often used to help students understand the properties of gasses in introductory chemistry and physics classes. This simulation is not optimized for a particular material or problem, so there are limits to the quantitative information that we can get from it. We use a Lennard-Jones potential (Figure \ref{fig:capped-lj}) in determining interaction forces for simplicity, but this only describes pairwise interactions. More sophisticated force fields are possible. Even with this simple simulation, we are able to generate a gas-like state and a solid-like state (Figure  \ref{fig:states-lj}), akin to those possible with other simple MD simulations that use a Lennard-Jones potential \citep{schroeder2015interactive,sweet2018facilitating}. 

\textbf{The Ephemeris experiment.} We downloaded raw ephemeris data for the inner planets of the solar system for the calendar year 2022 (1 day resolution). To do this we used the online interface provided by NASA's Horizons project. We used the Solar System Barycenter (SSB) for a coordinate center. In constructing our simulation, we used the simple gravitational well potential of $\frac{Gm_i m_j}{r_{ij}}$ and used SI units for the gravitational constant $G$, planet masses $m$, and durations of time. In exploring whether action minimization could reconstruct the inner planetary orbits, we perturbed only the paths of the inner planets and not that of the Sun. We considered a time duration of two months for this experiment because the orbits of Venus and Mercury cycle more rapidly than that of the Earth (qualitative visualizations of initial and final paths grew difficult as their orbits begin to extend over more than one cycle and overlap their tails).

\subsection{Additional quantum}
\label{appx:quantum}
\textbf{The principle of least action as a limiting case of the quantum path integral}. The principle of least action is just the limiting case of a path integral where the actions of paths considered are allowed to approach classical scales. In this regime, paths whose neighbors have variation in their actions contribute phases which are out of sync with one another. Summing contributions from these regions is akin to integrating a rapidly-oscillating sine function: every section with a positive (constructive) contribution is inevitably averaged out by an adjacent negative (destructive) contribution. As spatial scales increase, oscillations along any given path become more frequent and the net contribution of any arbitrary group of paths approaches zero. The sole exception occurs where groups of nearby paths have vanishing variations in action. These paths contribute similar phases and interfere constructively even at large spatial scales. \textit{From this point of view, the classical principle of least action is actually a description of the resonance modes of the quantum path integral}.

 \textbf{Additional discussion related to Figure \ref{fig:fig4}.} The quantum-classical resonance limit is depicted graphically in Figure \ref{fig:fig4}, where an identical quantum simulation is run at three different spatial scales. Notice that as we trend up in scale towards classicality, even small movements along off-diagonals result in a series of rapid oscillations. These are high-action paths, linear traversals of a particle between distant points in our space grid over a vanishing dt. Contributions from these paths will interfere destructively with one another in almost every trial. We see the direct result of this in the timed snapshots of each simulation; at a larger scale, the same initial wave packet will decohere slowly and exhibit only fine high-frequency interference structures compared to its lower-frequency and thus more-quantum counterparts. As we approach the classical limit, the entire wave packet will follow the classical path of least action for a particle bouncing off a wall with increasing fidelity.

\end{document}